\documentclass{fairmeta}

\ifdefined\pdfsuppressptexinfo\pdfsuppressptexinfo=15\fi
\ifdefined\pdfinfoomitdate\pdfinfoomitdate=1\fi
\ifdefined\pdftrailerid\pdftrailerid{}\fi

\usepackage{amsmath}
\usepackage{amssymb}
\usepackage{array}
\usepackage{longtable}
\usepackage{enumitem}
\usepackage{placeins}
\providecommand{\Description}[1]{}
\metabrandingtrue

\title{\LARGE CLEAR: Legible and Steerable Recommendation via Grounded Natural-Language Interests}
\hypersetup{
  pdftitle={CLEAR: Legible and Steerable Recommendation via Grounded Natural-Language Interests},
  pdfauthor={Qi Guo, Reza Mirghaderi, Iordanis Fostiropoulos, David Fan, Yuchen Liu, Shiliang Tang, Lukasz Korycki, Golnaz Ghasemiesfeh, Xinzhu Bei, Sasha Vorontsov, Hanchao Yu, Brendan Quinlivan, Khoi Pham, Jigar Gada, Hao Yu, Felix Li, Hang Gao, Boyu Fang, Yang Gao, Yale Thomas, Xiaomeng Jin, Avishek Dutta, Hong Hong, Janet Xia, Ethan Zhu, Ellie Pan, Xiang Shen, Jawwad Zakaria, Jinfeng Lin, Jonathan Rush, Ruixue Lian, Noah Kennedy, Brian Vanover, Jas Pyneni, Daniel Thiyagu, Apoorv Saxena, Sanyogeeta Lawande, Huayu Ding, Hima Cherukuru, Yang Liu, Chen Kovacs, Qunshu Zhang, Lizhu Zhang, Benyu Zhang, Kate Chen, Ashish Singal, Danny Kanoff, Zhiyu Liu, Jiayi Liu, Jianyu Wang, Fei Liu, Xiangjun Fan, Abhishek Kumar, Hong Yan},
  pdfsubject={An end-to-end recommendation system with an editable natural-language interest layer deployed at Meta},
  pdfkeywords={recommender systems, user profiles, transparency, user control, Meta, Instagram, Facebook}
}
\author[\dagger,*]{Qi Guo}
\author[\dagger,*]{Reza Mirghaderi}
\author[*]{Iordanis Fostiropoulos}
\author{David Fan}
\author{Yuchen Liu}
\author{Shiliang Tang}
\author{Lukasz Korycki}
\author{Golnaz Ghasemiesfeh}
\author{Xinzhu Bei}
\author{Sasha Vorontsov}
\author{Hanchao Yu}
\author{Brendan Quinlivan}
\author{Khoi Pham}
\author{Jigar Gada}
\author{Hao Yu}
\author{Felix Li}
\author{Hang Gao}
\author{Boyu Fang}
\author{Yang Gao}
\author{Yale Thomas}
\author{Xiaomeng Jin}
\author{Avishek Dutta}
\author{Hong Hong}
\author{Janet Xia}
\author{Ethan Zhu}
\author{Ellie Pan}
\author{Xiang Shen}
\author{Jawwad Zakaria}
\author{Jinfeng Lin}
\author{Jonathan Rush}
\author{Ruixue Lian}
\author{Noah Kennedy}
\author{Brian Vanover}
\author{Jas Pyneni}
\author{Daniel Thiyagu}
\author{Apoorv Saxena}
\author{Sanyogeeta Lawande}
\author{Huayu Ding}
\author{Hima Cherukuru}
\author{Yang Liu}
\author{Chen Kovacs}
\author{Qunshu Zhang}
\author{Lizhu Zhang}
\author{Benyu Zhang}
\author{Kate Chen}
\author{Ashish Singal}
\author{Danny Kanoff}
\author{Zhiyu Liu}
\author{Jiayi Liu}
\author{Jianyu Wang}
\author{Fei Liu}
\author{Xiangjun Fan}
\author{Abhishek Kumar}
\author{Hong Yan}
\affiliation{Meta}
\contribution[\dagger]{Joint first authors and technical leads}
\contribution[*]{Corresponding authors}

\abstract{Modern recommenders learn detailed preferences but keep them in representations that people cannot inspect or change. We present CLEAR (Controllable and Legible End-to-end Architecture for Recommendation), an end-to-end system deployed at Meta and built around an editable natural-language interest layer. CLEAR makes a ranked subset of inferred interests readable and stores accepted additions and suppressions as authoritative preference state. Behavioral grounding maps each inferred interest back to supporting events in the user's own history. The resulting interest object is also a serving address: its label and profile-conditioned rewrites drive lexical retrieval, its supporting events seed behavioral retrieval, its projection queries an embedding index, and its stable identity supports ranking, attribution, and exposure control. A fresh user declaration may precede behavioral support, so it enters through the same serving identity and initially draws candidates from topical retrieval. Ranking makes those candidates competitive, while a delivery controller observes attributable exposure and replenishes underexposed interests. Direct user review found that users rated about 88.5\% of displayed interests accurate. In an offline benchmark using a shared evaluation protocol, the Specialized 3B generator reached 0.3967 mean-user F1@10, compared with 0.3633 for Gemini 2.5 Pro and 0.1911 for the prior generator. For a newly declared interest, a user-randomized cumulative ablation measured same-day delivery losses of 8.7 percentage points for added-interest ranking terms, 11.1 for reserved top-slot insertion, and 13.7 for interest-conditioned candidate sourcing, conditional on the tested removal sequence. CLEAR components have operated in production for more than one year; the current system spans four surfaces across two platforms, Instagram and Facebook, and provides one interest layer that people can inspect, correct, and use to direct recommendations.}

\correspondence{Qi Guo at \email{qiguokq@meta.com}, Reza Mirghaderi at \email{rezam@meta.com}, Iordanis Fostiropoulos at \email{iordanis@meta.com}}
\date{October 24, 2026}

\begin{document}
\maketitle
\raggedbottom



\setlength{\textfloatsep}{9pt plus 2pt minus 2pt}

\section{Introduction}
\label{sec:intro}

Recommenders continually infer what a person may want, yet their operative representations are usually
latent and inaccessible to the person they describe. Item-level reactions can influence future
predictions, but they do not reveal what durable preference the system inferred or whether a correction
persisted. A useful control surface therefore needs two linked properties: the inferred interest state
must be \emph{legible}, and an accepted edit must be \emph{consequential}.

We call an interest layer \textbf{legible} when its natural-language entries let people assess whether
the system's inferences match their interests. Recommendations are \textbf{steerable through the layer}
when accepted edits become persistent preference state consumed by serving, and controlled changes to
that serving path measurably change later delivery. This definition separates seeing an interest from
proving that the recommender acts on it.

\begin{figure}[t!]
  \centering
  \includegraphics[width=\textwidth]{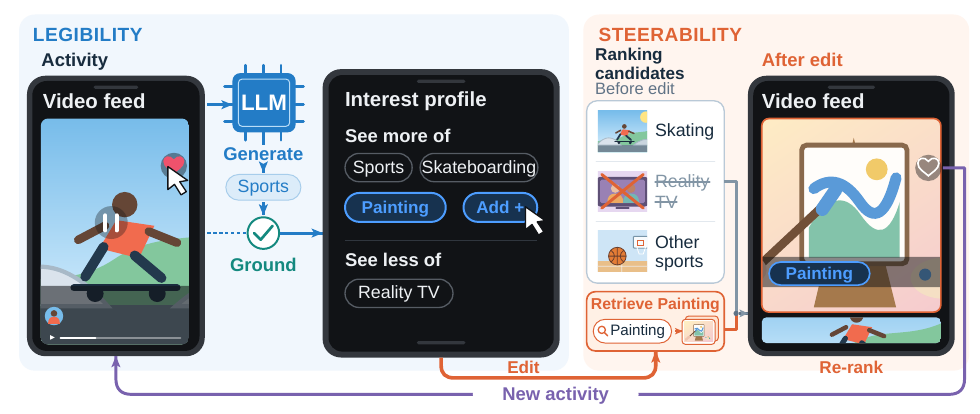}
  \Description{An activity feed is summarized as editable natural-language interests. A user adds Painting and suppresses Reality TV; topical retrieval expands the candidate pool, re-ranking changes the feed, and responses return to activity history.}
  \caption{\textbf{Generate, inspect, edit, retrieve, and re-rank through one interest layer.}
  An LLM proposes interests from recent activity, and behavioral grounding checks inferred entries before
  they enter the displayed profile. The user suppresses Reality TV and adds Painting. A fresh addition
  receives a stable serving identity immediately even when it has no prior support events; topical
  retrieval adds Painting candidates, and ranking makes them competitive in the next feed. Subsequent
  responses return through ordinary interaction history. Rankings are schematic.}
  \label{fig:main}
\end{figure}

CLEAR joins the read and write paths through one normalized interest identity (Figure~\ref{fig:main}).
It converts recent activity into natural-language candidates, behaviorally grounds inferred candidates,
exposes a ranked subset, and stores explicit additions and suppressions with precedence over later
inference. The same identity is read by retrieval, ranking, cross-source attribution, and delivery
control. It is an authoritative interest layer inside the recommender, not a claim that the system's
entire latent state is visible.

The system must bridge two kinds of meaning. For an inferred entry, grounding asks which of this user's
events support the phrase. For recommendation serving, the same interest object must identify where to
retrieve, which candidates are on topic, which delivered items count toward the interest, and when more
supply is needed. Grounding is therefore not confined to profile construction: it is the interface
between natural-language state and the recommendation cascade.

We evaluate the system in the same order. Direct review tests whether displayed entries are readable and
accurate. A randomized cumulative ablation then tests how much each serving layer contributes to delivery after
a positive edit. Supporting evaluations quantify grounding, incremental updates, and deployment efficiency.

\begin{figure}[t]
  \centering
  \includegraphics[width=0.60\textwidth]{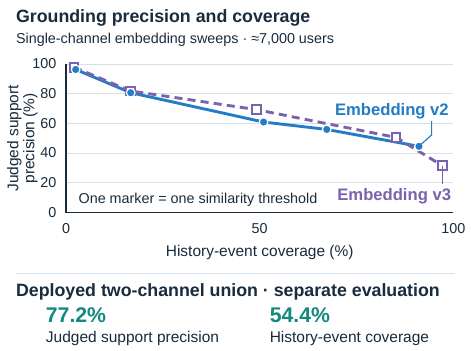}
  \Description{Two threshold sweeps trade judged support precision against history-event coverage. A separately evaluated deployed two-channel union reports 77.2 percent precision and 54.4 percent coverage.}
  \caption{\textbf{Grounding precision--coverage trade-off.} Curves show threshold sweeps for two
  single-channel embedding matchers. The deployed text-plus-media union is evaluated separately and
  summarized below the plot. Support precision concerns judged event--interest matches; coverage counts
  deduplicated history events. Table~\ref{tab:grounding} reports the selected operating-point results.}
  \label{fig:frontier}
\end{figure}

This paper contributes:
\begin{itemize}
  \item an editable interest-state contract that joins an inspectable natural-language profile to
  recommendation serving (Figure~\ref{fig:main});
  \item a grounding interface that turns inferred interests into personalized retrieval anchors and
  stable recommendation-side identities (Section~\ref{sec:grounding}, Figures~\ref{fig:write-detail}
  and~\ref{fig:frontier}, and
  Table~\ref{tab:grounding});
  \item randomized evidence that sourcing, placement, and ranking contribute to positive-edit delivery,
  and similarity-aware demotion reduces suppressed-interest reappearance
  (Tables~\ref{tab:main} and \ref{tab:stack-ablation}); and
  \item a deployed design whose incremental inference and shared annotation reduce serving cost while
  preserving profile quality (Tables~\ref{tab:incremental}--\ref{tab:efficiency}).
\end{itemize}

\section{Related Work}
\label{sec:related}

\textbf{Scrutable and editable profiles.}
Natural-language user profiles have been proposed as transparent, revisable interfaces to recommendation
\citep{Radlinski2022on,Ramos2024transparent}. Prior systems study simulated edits and reranking
\citep{Penaloza2024tears,Zhou2026steere}, direct edits to concept profiles
\citep{Mysore2023editab}, natural-language control requests \citep{Carroll2025ctrlre}, and preference
masking \citep{Liu2024filter}. CLEAR studies a complementary systems question: how one open-vocabulary
interest identity can persist from an inferred profile or user edit through candidate supply, ranking,
attribution, and exposure-aware delivery in an existing recommender.

Field and interface studies further examine how people reflect on, inspect, and revise interest summaries
\citep{Sun2025coauth,Wu2026rethin}. Stateful profile systems such as LIBER and RECAP update textual
representations as behavior evolves \citep{Zhu2024liber,Zhao2026recap}. CLEAR's contribution is the
deployed state contract that gives one editable identity persistent precedence and makes it consumable
through the recommendation cascade without replacing the incumbent rankers.

\textbf{Textual representations and deployed controls.}
Consumer controls commonly adjust service-defined topics, follows, or individual items
\citep{google2024interests,tiktok2025manage,ricks2022button}. LLM-derived textual user representations
also support recommendation \citep{Arora2026high,Wang2026llmbas,Yi2025recgpt,Xu2026shape,Zhu2024liber}.
CLEAR makes an inferred textual representation inspectable, gives explicit edits precedence, and
measures the recommendation components that consume those edits.

\textbf{Grounding language in recommendation.}
Attribution asks whether generated claims are supported by their sources
\citep{Rashkin2021measur,Min2023factsc}. For user interests, the source is the user's own interaction
history; GISTBench evaluates whether inferred interests can be verified against that evidence
\citep{fostiropoulos2026gistbench}. CLEAR operationalizes this connection. The support set filters and
orders inferred interests, while the interest's label, projection, anchors, and identity make the same
concept addressable to recommendation serving. The contribution is this conjunction with durable edit
precedence and measured production actuation, not a claim that editable profiles or any individual
retrieval mechanism is new.

\section{Editable Interest State}
\label{sec:state}

\subsection{The profile object}

In production, Meta's Biography service builds CLEAR's inferred interest profiles. Instagram's Your Algorithm and Tune Your Explore interfaces expose the interest layer for inspection and editing.

CLEAR represents interests at coarse, intermediate, and open-vocabulary levels. Coarser entries provide
stable organization; open-vocabulary entries provide the specificity needed for retrieval. A surface
displays a ranked subset of a larger served interest layer using natural-language labels rather than
model coordinates. This displayed subset is the legibility surface evaluated below.

Each served record binds a normalized interest identity to its source, supporting evidence when
available, query projection, shared annotations, and preference state. State can be inferred, explicitly
added or promoted, or explicitly removed or suppressed. Explicit state is stored separately from model
output and takes precedence until the user changes it. Inferred interests may expire when their support
decays; accepted user preferences persist across inference refreshes. Table~\ref{tab:state-contract}
summarizes the shared state and action contract.

\subsection{Profile construction}

Construction runs asynchronously over a bounded representation of recent activity. History selection
favors topical coverage rather than using recency alone, and related events are grouped before the
history and contextual signals enter the language model. The generator proposes candidate interests.
Generation, behavioral grounding, policy filtering, deduplication, and display ranking are separate
stages so that a fluent candidate is not automatically treated as supported or shown.

New candidates merge with the prior inferred profile. Removing old history may remove behavioral support;
recency changes the ordering of entries that remain supported. Explicit additions and suppressions are
kept outside this inferred-state merge, which prevents later inference from silently undoing a user's
correction.

\subsection{Incremental state and shared annotation}

Incremental inference reuses the prior profile and processes new activity instead of regenerating from
the full history. The bounded profile is written to a serving cache. User-independent annotations, such
as normalized wording and sensitivity, attach once to each distinct interest string and are reused across
user records. User-specific membership, support events, and explicit preference state are never shared.

\section{Grounding Natural-Language Interests}
\label{sec:grounding}

\subsection{Behavioral grounding for inferred interests}

An inferred phrase is useful only if the system can relate it to what the user actually did. CLEAR
projects the phrase into the recommender's co-engagement space and performs reverse lookup over the
user's own history. A content-understanding channel and a co-engagement-trained media channel retrieve
candidate support events; their deduplicated union forms the interest's support set.

The support set has three profile-side consumers. Its size filters inferred candidates with insufficient
evidence. Its overlap reveals personalized near-duplicates: two phrases may be redundant for one user
when the same activity supports both. Event actions and recency supply an evidence-weighted ordering
signal. Grounding therefore verifies behavioral support, while direct user review separately tests
whether the person endorses the resulting description.

Appendix Figure~\ref{fig:grounding-detail} shows how the matched events are reused by profile filtering,
deduplication, ordering, and coverage accounting.

A user declaration follows a different contract. A fresh addition can precede any matching history and
therefore bypasses the inferred-candidate support requirement. It still receives the same normalized
identity, query projection, policy checks, and topical matching path. If relevant prior activity exists,
it can supply personalized anchors immediately; later responses enter ordinary history and can support
subsequent inference. This distinction avoids treating a cold-start declaration as evidence about past
behavior.

\subsection{Grounding as a recommendation-side address}

The grounded interest object exposes four fields to the recommendation cascade:
\begin{itemize}
  \item the \emph{label}, expanded with profile-conditioned rewrites, becomes a personalized lexical query;
  \item the \emph{support events} seed trigger-to-item retrieval in the user's own co-engagement region;
  \item the \emph{query projection} retrieves nearby items from an approximate-nearest-neighbor index; and
  \item the stable \emph{user--interest identity} keys cached supply, attribution, and exposure accounting.
\end{itemize}

Behavioral retrieval uses a conjunction rather than either signal alone: a candidate must lie near the
user's own support events and match the interest under content-understanding features. This excludes a
globally on-topic item unrelated to the user's behavioral neighborhood and a behaviorally adjacent item
that is off topic. The lexical and embedding paths provide initial supply for a novel declaration; the
support-set path personalizes that supply when behavioral evidence exists. A separate generative source
can condition on the same user--interest key.

\begin{figure}[t!]
  \centering
  \includegraphics[width=\textwidth]{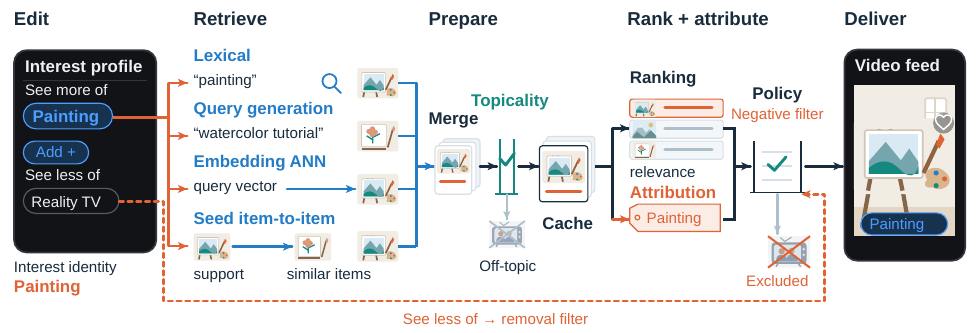}
  \Description{A stable interest object supplies lexical, profile-expanded, behavioral-anchor, embedding, and generative retrieval before topical filtering, caching, ranking, attribution, and delivery.}
  \caption{\textbf{Grounding gives one editable interest four retrieval addresses and a stable serving
  identity.} The display label becomes a lexical query, profile-conditioned rewrites expand that query,
  inferred interests retain behavioral support for trigger-to-item retrieval, the interest projection
  becomes an embedding query, and the user--interest identity can condition generative retrieval and
  cached supply. Candidates merge, pass a hard topicality gate, and enter ranking. Attribution preserves
  the same interest identity for delivery control; explicit negative state remains in the policy path.}
  \label{fig:write-detail}
\end{figure}

Stable identity also closes the recommendation-side loop. Ordinary and dedicated-source candidates can
be attributed to the same declared interest, so the controller observes total recent exposure rather
than only what its own retriever injected. The same identity then determines whether the system should
request more topical candidates or reduce repeated supply. Figure~\ref{fig:write-detail} shows the
resulting path from the editable interest object through retrieval, topicality, ranking, attribution, and
delivery.

\section{From an Edit to Delivery}
\label{sec:serving}

\subsection{Write contract and candidate supply}

At Meta, the declared-interest serving path is called Carrera.

User text is parsed and normalized before storage. Positive actions add or promote an identity; negative
actions remove or suppress it. The same integrity gates apply to inferred and user-entered interests.
An accepted edit immediately updates the served layer, and explicit state remains authoritative across
later inference.

Four interest-conditioned retrieval mechanisms emit tagged candidates: lexical retrieval,
trigger-to-item retrieval, embedding retrieval, and generative retrieval. An asynchronous path
precomputes per-user, per-interest candidates; a request-time path handles fresh declarations and stale
cached supply. Candidates merge with the ordinary pool before ranking. Candidate supply therefore changes which
items can compete, rather than merely reordering a fixed pool.

\subsection{Ranking, topical matching, and attribution}

Early- and late-stage ranking apply temporary declaration-conditioned terms so candidates for a new
interest can compete before sufficient behavior accumulates. A small number of top-ranked positions can
also be reserved for candidates matching a recent declaration, guaranteeing placement rather than only
competitiveness. Negative state is read by a serving-time removal path. Both operations act on the
normalized identity, not only its displayed text.

A compact learned scorer estimates whether an item matches an interest from semantic and
content-understanding features. Retrieval uses a hard topicality gate; ranking can use the score softly.
The same scorer also attributes ordinary recommender candidates. Source identity answers where a
candidate came from, while topical attribution answers which interest it matches. These are intentionally
separate: an ordinary candidate can satisfy a declared interest without coming from dedicated retrieval.
This item--interest scorer is distinct from both the semantic judge used to compare generated profiles
and the history-event matcher used to ground inferred interests.

\subsection{Exposure-aware delivery}
\label{sec:control}

Naively boosting every match can flood one topic. CLEAR separates topical relevance from how much recent
exposure that user--interest pair has received. Attribution runs before the controller, allowing exposure
accounting to include both ordinary and dedicated-source matches.

Fireball maintains a decayed real-time counter for realized per-interest exposure (RPE). Higher exposure dampens further promotion; as exposure
decays, the interest becomes eligible for renewed sourcing. Replenishment dynamically reserves recent
declarations and prioritizes eligible interests with least recent realized exposure, subject to a
recent-view threshold. Replenishment changes whether an interest receives fresh supply, whereas ranking
changes the competitiveness of candidates already present.

The controller never writes model parameters. Delivered items and user responses enter the
recommender's ordinary learning loop, which may later improve organic supply. The controller's faster
loop only retrieves, ranks, attributes, and regulates current delivery.

Appendix Figure~\ref{fig:controller-detail} summarizes the fast exposure loop and slower learning loop.

Together these paths form an import, teach, harvest, and re-teach design loop. A fresh declaration first
imports topical candidates because the base recommender may have little evidence for it. User responses
to those candidates become ordinary behavioral signal. As the recommender later supplies organic
candidates attributed to the same interest, the controller can count and use that personalized supply
instead of relying only on imported candidates. If attributable exposure later declines, replenishment
reopens topical sourcing. This sequence explains the controller design, and the selected studies
evaluate its supply, ranking, replenishment, and reception components.

\section{Evaluation}
\label{sec:evaluation}

Bracketed uncertainty intervals denote 95\% confidence intervals.

\paragraph{Three distinct measurement layers.}
Generator evaluation compares natural-language interests with a user-validated interest reference using
a semantic judge. Behavioral-grounding evaluation uses a Gemini LLM judge on retrieved
history-event/inferred-interest pairs; \emph{Vital} or \emph{Reasonable} counts as supported. Serving
evaluation instead scores delivered-item/declared-interest pairs. On 137 human-labelled event-interest
pairs, grounding-judge variants agreed with human labels at 93--98\%, with Cohen's $\kappa$ of
0.86--0.95. Direct review remains a user judgment over displayed interests.
Table~\ref{tab:experiment-contracts} collects the study-level comparison and analysis contracts.

\subsection{Legibility and profile quality}

We operationalize legibility as displayed-profile accuracy: a person can inspect a natural-language
entry and judge whether it reflects their interests. Direct review pools 7,261 interest ratings from
393 users. Rejection reasons are analyzed separately to distinguish exposure from endorsement.

The current offline generator benchmark uses a shared evaluation protocol for the Specialized 3B
generator, Gemini 2.5 Pro, and the prior production generator. All receive the same history format and
are scored against the same user-validated reference by the same semantic judge. Users validate the
reference by confirming proposed interests and supplying omissions. F1@10, P@10, and R@10 evaluate
the first ten emitted interests; F1@all evaluates the complete generated profile.
Table~\ref{tab:main} reports mean-user F1@10 first, followed by precision, recall, and whole-profile F1.
Generation is evaluated before behavioral grounding and display selection.

The Specialized 3B generator is trained with a task-specific curriculum aligned to the fixed top-ten
interest-generation objective. At that operating point, it reaches 0.8067 P@10, compared with 0.7982 for
Gemini 2.5 Pro.

\subsection{Grounding}

The support-filter study compares retained inferred profiles with an LLM-generated
whole-profile reference. Precision measures the fraction of retained interests matching the reference;
recall measures the fraction of reference interests retained. It is distinct from F1@10.

The event-level grounding study judges whether retrieved history events support an interest.
Vital denotes strong relevance to the post's most fine-grained information; Reasonable denotes broad
relevance to its general topic. Both receive equal positive credit in this history-matcher evaluation.
Support precision is the fraction of event--interest matches judged relevant. History-event coverage is the
fraction of a user's history matched to at least one interest, with an event counted once even when
several interests retrieve it.

\subsection{Steerability}
\label{sec:steerability-eval}

We evaluate Instagram's declared-interest serving at two levels. At the system level we report three production measures. \emph{Same-day responsiveness} is defined per user as the share of that user's newly
declared interests that received at least one attributed matching video later the same day, averaged over
users who subsequently viewed more than ten videos. \emph{Prevalent declared-interest load} is the share
of eligible users holding a declared interest for whom content attributed to declared interests reaches
at least five percent of their recommendations. \emph{Topicality} is evaluated by a judge independent of
serving. Twice weekly, approximately 8,000 (video, interest) pairs, about 10,500 impressions, are sampled
from impressions attributed to explicit declared interests, and a multimodal judge consuming the video
content directly, rather than captions or metadata, adjudicates whether the item matches the interest it
was served against. The reported value is the impression-weighted share of matching pairs. This judge is
not the serving topicality scorer, so it does not inherit the admission criterion it evaluates. The judge
is itself validated against human labels before use and re-validated across rounds as prompts and models
change: on a human-labelled golden set of 137 (item, interest) pairs, deployed judge variants agree with
human labels at 93--98\% (Cohen's $\kappa=0.86$--$0.95$) and with one another at 95--96\%
($\kappa=0.89$--$0.92$). The calibration set oversamples edge cases and is therefore not
production-representative.

At the component level, a five-arm stack ablation removes serving layers cumulatively from the production
configuration: added-interest ranking boosts, then forced top-slot insertion, then negative filtering and
suppression, then interest-conditioned candidate sourcing. Differences between consecutive arms give the
marginal delivery loss at each removal step; the outer-arm contrast compares all layers removed with the
production control. The endpoint is same-day attributed matching delivery.

A separate randomized study compares dynamic topic selection with the prior age-limited policy, measuring
the share of active interests receiving a match watched for at least three seconds and user days with
explicit-interest delivery.

A six-day user-randomized study compares real-time per-interest accounting and recalibrated volume
dampening with the prior daily-accounting configuration. Its endpoint is the recorded flooding-days
metric; lower is better.

\section{Results}
\label{sec:results}

\begin{table}[t!]
\centering
\small
\setlength{\tabcolsep}{3.8pt}
\renewcommand{\arraystretch}{1.12}
\caption{\textbf{Legibility and steerability.}
Direct review and system-level rows report measured levels; replenishment and volume regulation report relative changes.
Direct review is descriptive, and the generator comparison is an offline benchmark using a shared history format,
user-validated reference, and semantic judge. Responsiveness and prevalence are one-week Instagram production averages; topicality is the most recent twice-weekly judge run. Responsiveness is attributed matching
delivery on the day of the edit; topicality is scored by a judge independent of the serving gate and
validated against human labels (Section~\ref{sec:steerability-eval}). Replenishment is a separate
user-randomized comparison against the prior age-limited policy. Volume regulation compares real-time
per-interest accounting and recalibrated dampening with prior daily accounting. Component-level contrasts appear in
Table~\ref{tab:stack-ablation}.}
\label{tab:main}
\begin{tabular}{@{}>{\raggedright\arraybackslash}p{0.24\textwidth}
    >{\raggedright\arraybackslash}p{0.47\textwidth}
    >{\raggedright\arraybackslash}p{0.24\textwidth}@{}}
\toprule
\textbf{Study} & \textbf{Metric} & \textbf{Result} \\
\midrule
\multicolumn{3}{@{}l}{\textbf{A. Legibility and profile quality}} \\*
\addlinespace
Direct review & Displayed interests rated accurate & $\approx$88.5\% \\
Specialized 3B generator & Mean-user F1@10 / P@10 / R@10 / F1@all & 0.3967 / 0.8067 / 0.2901 / 0.5493 \\
Gemini 2.5 Pro & Mean-user F1@10 / P@10 / R@10 / F1@all & 0.3633 / 0.7982 / 0.2568 / 0.7051 \\
Prior production generator & Mean-user F1@10 / P@10 / R@10 / F1@all & 0.1911 / 0.5692 / 0.1281 / 0.4281 \\
\midrule
\multicolumn{3}{@{}l}{\textbf{B. Steerability}} \\*
\addlinespace
Same-day responsiveness & Share of a user's newly declared interests receiving a matching video the same day & 79.6\% \\
Prevalent declared-interest load & Users whose declared-interest content is at least 5\% of their recommendations & 82.6\% \\
Topicality & Interest-matched impressions judged on-topic by an independent multimodal judge & 87.2\% \\
Replenishment & 3s-engaged interest responsiveness / user days with explicit-interest delivery &
+8.34\% [+8.07\%, +8.61\%] / +2.43\% [+2.24\%, +2.63\%] \\
Volume regulation & Recorded flooding-days metric versus prior daily accounting &
15.4\% lower [13.4\%, 17.4\%] \\
\bottomrule
\end{tabular}
\end{table}

\subsection{Displayed interests are accurate; the specialized generator improves F1@10}

Across 7,261 displayed interests rated by 393 users, approximately 88.5\% were rated accurate
(Table~\ref{tab:main}). In the separate
offline benchmark using a shared evaluation protocol, the Specialized 3B generator reached 0.3967 mean-user F1@10, compared with 0.3633
for Gemini 2.5 Pro and 0.1911 for the prior generator.
F1@10 evaluates the fixed top-ten generation objective. Gemini 2.5 Pro achieved higher
whole-profile F1@all: 0.7051 versus 0.5493 for the Specialized 3B generator.
The direct review evaluates the shipped display;
the benchmark isolates generation before grounding and display selection.

Rejected entries also show why support and endorsement differ. Among rejected interests, 41\% reflected
content the user had viewed without personal interest, 29\% reflected transient interests, and 21\%
reflected no personal interest (Table~\ref{tab:survey-detail}). A readable profile exposes these
mismatches so the user can correct them.

\subsection{Stricter grounding trades reference recall for whole-profile precision}

Increasing the minimum support from zero to three events moved whole-profile precision from 69.9\% to
77.4\% and reference recall from 46.2\% to 28.7\%, the expected precision--recall trade-off from stricter
grounding. The threshold sensitivity analysis selects the three-event point to prioritize
whole-profile precision for displayed interests (Table~\ref{tab:grounding}). In the separate event-level
evaluation, the deployed text-plus-media matcher reached 77.2\% judged support precision and matched
54.4\% of deduplicated history events to at least one retained interest.

\subsection{The deployed system responds to edits and delivers on-topic content}

Before isolating components, we report how the deployed system behaves as a whole in Instagram production. On the day a user declared new interests, the average user received a matching video
for 79.6\% of those interests later the same day. For 82.6\% of eligible users holding a declared
interest, content attributed to declared interests made up at least 5\% of their recommendations, so the
declared-interest load is prevalent rather than token for the large majority of users. An independent
multimodal LLM judge rated 87.2\% of sampled interest-matched impressions as on-topic for the interest
they were served against, over approximately 10,500 impressions drawn from 8,000 video--interest pairs
per run.

These system-level measures (Section B of Table~\ref{tab:main}) answer a different question than the component studies
that follow. Responsiveness reports the absolute rate at which a declaration is honored, rather than a
relative change against an arm with the component removed, and topicality is scored independently of the
serving gate.

\textbf{Reception by attribution status.}
Among viewers holding at least one declared interest, recommendations attributed to a declared interest
were liked at 11.55\% against 9.90\% for other recommendations shown to that cohort, a difference of
1.65 percentage points, and the ratio of positive to negative content-control taps was 6.44 against 2.95
(2.2 times). The attributed
subset drew 1.6 percentage points more three-second skips and 10\% less dwell per impression. These are
descriptive cohort-level differences between delivered content subsets (Table~\ref{tab:reception}).

Attributed recommendations receive more explicit positive feedback, while their higher skip rate and
lower dwell show mixed overall reception.

\begin{table}[t!]
\centering
\small
\setlength{\tabcolsep}{3.8pt}
\renewcommand{\arraystretch}{1.12}
\caption{\textbf{Descriptive reception against recommendation and stated-intent references.}
Among viewers holding at least one declared interest, the first block compares attributed and other
recommendations on Instagram Reels. The second uses videos viewed through Instagram Search as a cross-surface stated-intent reference.}
\label{tab:reception}
\begin{tabular}{@{}>{\raggedright\arraybackslash}p{0.33\textwidth}
    >{\raggedright\arraybackslash}p{0.61\textwidth}@{}}
\toprule
\textbf{Metric} & \textbf{Observed comparison} \\
\midrule
\multicolumn{2}{@{}l}{\textbf{Same-surface reference: attributed vs. other recommended}} \\*
Like rate & 11.55\% attributed vs. 9.90\% other recommended; +1.65 pp \\
Positive-to-negative content-control tap ratio & 6.44 attributed vs. 2.95 other recommended; 2.2$\times$ \\
Three-second skip rate & +1.6 pp on attributed recommendations \\
Dwell per impression & $-$10\% on attributed recommendations \\
\midrule
\multicolumn{2}{@{}l}{\textbf{Cross-surface stated-intent reference: attributed vs. search-result videos}} \\*
Like rate & 1.9$\times$ on attributed recommendations \\
Three-second skip rate & $-$8.4 pp on attributed recommendations \\
Dwell per impression & +26\% on attributed recommendations \\
\bottomrule
\end{tabular}
\end{table}

\subsection{Component studies isolate supply, competition, and continuity}

\begin{table}[t!]
\centering
\small
\setlength{\tabcolsep}{5pt}
\renewcommand{\arraystretch}{1.12}
\caption{\textbf{Controlled serving-component effects for positive and negative edits.}
Panel A is a five-arm user-randomized cumulative ablation of Instagram's positive-addition delivery layers. Component rows report consecutive-arm delivery losses in the tested removal order; the full-stack
row reports the outer-arm contrast. All reported randomized
serving-component comparisons have $p<0.01$.
The endpoint uses a slightly narrower definition than the same-day responsiveness in Table~\ref{tab:main}
and is read as an approximation of it. Because the arms are nested, the decomposition is specific to this
removal order and is not an order-free attribution. Negative filtering and suppression produced no
measurable change on this positive-addition endpoint and are omitted from Panel A. Panel B is a separate
user-randomized suppression experiment; lower suppressed-interest reappearance is better.}
\label{tab:stack-ablation}
\begin{tabular}{@{}>{\raggedright\arraybackslash}p{0.70\textwidth}
    >{\raggedright\arraybackslash}p{0.24\textwidth}@{}}
\toprule
\multicolumn{2}{@{}l}{\textbf{A. Positive addition}} \\*
\textbf{Layer removed} & \textbf{Loss in same-day delivery} \\
\midrule
Interest-conditioned candidate sourcing & 13.7 pp \\
Forced top-slot insertion & 11.1 pp \\
Added-interest ranking boosts & 8.7 pp \\
\textbf{All layers removed} & \textbf{33.5 pp} \\
\midrule
\multicolumn{2}{@{}l}{\textbf{B. Negative suppression}} \\*
\textbf{Treatment} & \textbf{Change in same-day suppressed-interest reappearance} \\
Similarity-aware demotion beyond the production filter & 11.8\% lower \\
\bottomrule
\end{tabular}
\end{table}

A five-arm stack ablation isolates each serving layer by removing them cumulatively from the production
configuration (Table~\ref{tab:stack-ablation}). With no layer enabled, a substantial share of newly
declared interests still receives a matching video the same day, because the recommender sometimes serves
on-topic content without being directed to; enabling the full stack adds 33.5 percentage points on top of
that baseline. Interest-conditioned candidate sourcing contributes the largest single share (13.7 pp),
followed by forced top-slot insertion (11.1 pp) and added-interest ranking boosts (8.7 pp). Supply,
placement, and competition therefore each contribute materially in the tested removal order, and no
single layer accounts for a majority of the effect. Negative filtering and suppression produced no measurable change on this endpoint, which is
consistent with their role: they remove unwanted content rather than deliver requested content.

Negative edits use a separate randomized study. Similarity-aware demotion extends the production filter
to ordinary recommendations and reduces reappearance of suppressed interests
(Table~\ref{tab:stack-ablation}; Appendix~\ref{app:serving-experiments}).

Separately, replenishment increased the share of active interests receiving a match watched for at least
three seconds by
8.34\% [8.07\%, 8.61\%] relative to the prior age-limited policy; Table~\ref{tab:main} and
Appendix~\ref{app:serving-experiments} report the companion responsiveness result.

\section{Deployment and Efficiency}
\label{sec:deployment}

CLEAR components have operated in production for more than one year; the current system spans four surfaces across two platforms, Instagram and Facebook. History compression uses 39.5\% fewer
prompt tokens, history budgeting reduces estimated accelerator need by 38.9\%, shared annotation removes
approximately 98\% of logical annotation requests, and the updated annotation pipeline raises throughput
per accelerator by 145.5\% (Table~\ref{tab:efficiency}). These are separate levers with separate
accounting baselines.

Incremental merging modestly improved whole-profile precision, recall, and F1 over full regeneration in
the 1,000-user-per-condition offline comparison ($p<0.01$); the incremental-inference capacity estimate separately requires
35\% fewer accelerators than the preceding history-budget configuration (Table~\ref{tab:incremental}).
A separate historical maximum-history input benchmark reports approximately 216 times the
normalized model-inference cost for Gemini 2.5 Pro relative to its 3B reference
(Table~\ref{tab:model-cost}; Figure~\ref{fig:cost}). Parallelizing request-time attribution with ranking reduced chunk-0 server
latency by approximately 5 ms (Table~\ref{tab:efficiency}).

\section{Discussion}
\label{sec:discussion}

\textbf{A shared interest identity aligns transparency and control.}
The central design choice is not to expose the recommender's entire internal state. CLEAR exposes a
ranked subset of an authoritative interest layer and makes the same normalized identities available to
serving. A person can inspect an inference, an explicit correction takes precedence, and later candidates
and impressions remain attributable to the corrected state.

\textbf{Grounding personalizes more than the profile.}
Behavioral support filters inferred interests, but its recommendation-side role is equally important.
Support events place retrieval in the user's own co-engagement region, profile-conditioned rewrites
personalize lexical search, conjunction-based admission combines behavioral proximity with topical
evidence, and stable identity lets ordinary and dedicated delivery share one exposure account. A novel
declaration begins without claiming past support, then uses the same address as evidence and organic
supply accumulate.

\textbf{Steerability uses supply, placement, and competition.}
The tested removal sequence assigns material contributions to sourcing (13.7 pp), reserved placement
(11.1 pp), and ranking competition (8.7 pp). Separately, replenishment raises the share of active interests receiving a match
watched for at least three seconds by 8.34\% [8.07\%, 8.61\%] (Table~\ref{tab:main}).

\textbf{Exposure accounting limits over-service of a single interest.}
The controller exists because naively promoting every topical match concentrates delivery on one
interest. A six-day user-randomized comparison of real-time per-interest accounting and recalibrated volume
dampening against the prior daily-accounting configuration reduced the recorded flooding-days metric
by 15.4\% [13.4\%, 17.4\%] (Table~\ref{tab:main}).

\textbf{Reception relative to explicit search.}
Table~\ref{tab:reception} contrasts search and same-surface references: dwell and skip favor
declared-interest content against search, but favor the cohort's other recommendations within the surface.

\textbf{Operational lessons from deployment.}
The production system separates four responsibilities that are easy to conflate: candidate supply,
placement, competition in ranking, and continuity through replenishment. The stack ablation isolates the
first three. The two-speed architecture also separates immediate control from ordinary recommender
learning: the controller can act on an accepted interest without retraining the base model, while later
user responses improve organic personalization through the existing learning loop. Topical gating and
periodic grounding keep candidate feedback tied to the same stable interest identity across release
cycles and surfaces.

\section{Conclusion}

CLEAR makes natural-language interests inspectable and operational. Grounding supplies support and
retrieval anchors; a shared identity carries edits through serving and exposure control. Direct review
establishes displayed-interest accuracy, while randomized studies establish actuation and continuity.
People can inspect, correct, and direct an existing recommender.

\FloatBarrier
\label{page:last-main}
\clearpage
\label{page:references}

\clearpage
\beginappendix
\label{page:appendix-start}

\setcounter{table}{0}
\renewcommand{\thetable}{A\arabic{table}}
\normalsize
\setlength{\LTpre}{2pt}
\setlength{\LTpost}{2pt}
\captionsetup{font=small,skip=2pt}

\section{Supporting Evaluations and Results}
\label{app:construction}

Tables~\ref{tab:survey-detail}--\ref{tab:incremental} report profile and grounding evaluations;
Tables~\ref{tab:model-cost}--\ref{tab:efficiency} report cost and deployment results.
Tables~\ref{tab:state-contract}--\ref{tab:experiment-contracts} define state and study contracts.
Appendix~\ref{app:serving-experiments} details serving experiments, and Appendix~\ref{app:steering}
shows the controller and grounding diagrams (Figures~\ref{fig:controller-detail}--\ref{fig:grounding-detail}). Each table defines its own
comparison or accounting baseline; studies with different references, populations, or costs are not
combined.

\subsection{Why corrections are necessary}

The rejection categories distinguish behavioral exposure from interests a person wants the system to
retain. They are shares of rejected displayed entries, not shares of all displayed interests.

\begingroup
\small
\setlength{\tabcolsep}{4pt}
\renewcommand{\arraystretch}{1.10}
\begin{longtable}{@{}>{\raggedright\arraybackslash}p{0.62\textwidth}
    >{\raggedright\arraybackslash}p{0.33\textwidth}@{}}
\caption{\textbf{Why profile corrections are necessary.} Reasons users gave for rejecting a displayed
interest. Categories sum to approximately 100\% after rounding.}
\label{tab:survey-detail}\\
\toprule
\textbf{Finding} & \textbf{Result} \\
\midrule
\endfirsthead
\caption[]{Why profile corrections are necessary (continued).}\\
\toprule
\textbf{Finding} & \textbf{Result} \\
\midrule
\endhead
\bottomrule
\endlastfoot
Viewed without personal interest & 41\% of rejections \\
Transient interest & 29\% of rejections \\
No personal interest & 21\% of rejections \\
Other responses & $\approx$9\% of rejections \\
\end{longtable}
\endgroup

\subsection{Grounding quality and coverage}

The support-count filter and the event matcher evaluate different stages. The first scores retained
profiles against an LLM-generated reference. The second judges whether retrieved history events support an
interest and measures how much history the support sets cover.

\begingroup
\small
\setlength{\tabcolsep}{4pt}
\renewcommand{\arraystretch}{1.10}
\begin{longtable}{@{}>{\raggedright\arraybackslash}p{0.20\textwidth}
    >{\raggedright\arraybackslash}p{0.28\textwidth}
    >{\raggedright\arraybackslash}p{0.32\textwidth}
    >{\raggedright\arraybackslash}p{0.14\textwidth}@{}}
\caption{\textbf{Grounding quality and coverage.} Profile filtering forms a whole-profile
precision--recall frontier: increasingly strict support retains fewer interests while raising precision.
The selected three-event point prioritizes whole-profile precision for displayed interests. Event grounding
separately evaluates retrieved history-event and inferred-interest pairs.}
\label{tab:grounding}\\
\toprule
\textbf{Evaluation} & \textbf{Configuration} & \textbf{Metric} & \textbf{Result} \\
\midrule
\endfirsthead
\caption[]{Grounding quality and coverage (continued).}\\
\toprule
\textbf{Evaluation} & \textbf{Configuration} & \textbf{Metric} & \textbf{Result} \\
\midrule
\endhead
\bottomrule
\endlastfoot
\multicolumn{4}{@{}l}{\textbf{Profile filtering frontier: increasingly strict support}} \\*
Profile filtering & Unfiltered candidates & Precision & 69.9\% \\
Profile filtering & Unfiltered candidates & Reference recall & 46.2\% \\
Profile filtering & At least one supporting event & Precision & 74.5\% \\
Profile filtering & At least one supporting event & Reference recall & 41.5\% \\
Profile filtering & At least three supporting events (selected) & Precision & 77.4\% \\
Profile filtering & At least three supporting events (selected) & Reference recall & 28.7\% \\
\midrule
\multicolumn{4}{@{}l}{\textbf{Event grounding over the user's history}} \\*
Event grounding & Text + media matcher & Gemini-judged event--interest support precision & 77.2\% \\
Event grounding & Text + media matcher & History events matched to at least one interest & 54.4\% \\
\end{longtable}
\endgroup

\subsection{Incremental profile updates}

Incremental merging incorporates new activity into the retained profile instead of regenerating from
the full history. The capacity estimate uses the same deployed incremental-inference configuration.

\begingroup
\small
\setlength{\tabcolsep}{4pt}
\renewcommand{\arraystretch}{1.10}
\begin{longtable}{@{}>{\raggedright\arraybackslash}p{0.34\textwidth}
    >{\raggedright\arraybackslash}p{0.42\textwidth}
    >{\raggedright\arraybackslash}p{0.19\textwidth}@{}}
\caption{\textbf{Incremental profile updates.} Full regeneration and incremental merging use the
same fixed reference; whole-profile precision, recall, and F1 are reported, with $p<0.01$ for the
incremental comparison. Capacity reduction is measured against the preceding history-budget configuration.}
\label{tab:incremental}\\
\toprule
\textbf{Configuration} & \textbf{Metric} & \textbf{Result} \\
\midrule
\endfirsthead
\caption[]{Incremental profile updates (continued).}\\
\toprule
\textbf{Configuration} & \textbf{Metric} & \textbf{Result} \\
\midrule
\endhead
\bottomrule
\endlastfoot
Full regeneration (1,000 users) & Whole-profile precision / recall / F1 & 0.6375 / 0.1101 / 0.1819 \\
Incremental merging (1,000 users) & Whole-profile precision / recall / F1 & 0.6501 / 0.1135 / 0.1867 \\
Incremental inference versus preceding history budget & Estimated accelerator requirement & 35\% lower \\
\end{longtable}
\endgroup

\subsection{F1@all, F1@10, and inference cost}

The comparison reports F1@all and mean-user F1@10 together. The separate historical cost benchmark
uses a common maximum-history input basis and a 3B reference normalized to 1$\times$.
It accounts for model inference and excludes retrieval, annotation, and recommendation serving.

\begingroup
\small
\setlength{\tabcolsep}{4pt}
\renewcommand{\arraystretch}{1.10}
\begin{longtable}{@{}>{\raggedright\arraybackslash}p{0.28\textwidth}
    >{\raggedright\arraybackslash}p{0.20\textwidth}
    >{\raggedright\arraybackslash}p{0.20\textwidth}
    >{\raggedright\arraybackslash}p{0.22\textwidth}@{}}
\caption{\textbf{F1@all, F1@10, and inference cost.} Costs use the historical common maximum-history input basis
with the 3B cost reference normalized to 1$\times$.}
\label{tab:model-cost}\\
\toprule
\textbf{Model} & \textbf{F1@all} & \textbf{F1@10} & \textbf{Relative inference cost} \\
\midrule
\endfirsthead
\caption[]{F1@all, F1@10, and inference cost (continued).}\\
\toprule
\textbf{Model} & \textbf{F1@all} & \textbf{F1@10} & \textbf{Relative inference cost} \\
\midrule
\endhead
\bottomrule
\endlastfoot
Specialized 3B & 0.5493 & 0.3967 & 1$\times$ \\
Gemini 2.5 Pro & 0.7051 & 0.3633 & 216$\times$ \\
\end{longtable}
\endgroup

\begin{center}
\begin{minipage}{0.62\textwidth}
  \centering
  \includegraphics[width=\textwidth]{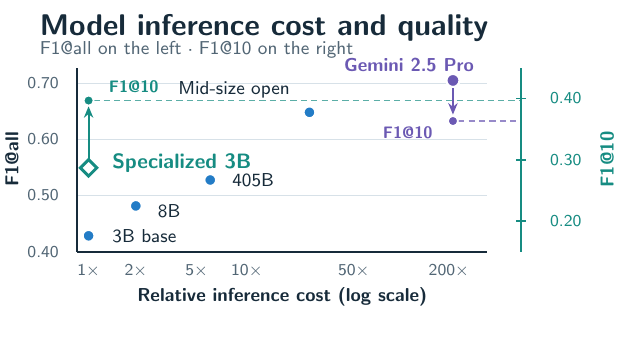}
  \captionof{figure}{\textbf{Model inference cost and quality.} Blue points use the left F1@all axis.
  Specialized 3B and Gemini 2.5 Pro connect vertically to their F1@10 values on the parallel right
  axis. Horizontal guides terminate at the right-axis values. Cost uses the shared log-scale x-axis.}
  \label{fig:cost}
\end{minipage}
\end{center}

\subsection{Deployment efficiency}

These rows evaluate distinct efficiency levers and are not multiplied together. Capacity entries are
estimates; token, request, and throughput entries follow their stated accounting baselines.

\begingroup
\small
\setlength{\tabcolsep}{4pt}
\renewcommand{\arraystretch}{1.10}
\begin{longtable}{@{}>{\raggedright\arraybackslash}p{0.30\textwidth}
    >{\raggedright\arraybackslash}p{0.46\textwidth}
    >{\raggedright\arraybackslash}p{0.19\textwidth}@{}}
\caption{\textbf{Deployment efficiency and breadth.} Each row has its own baseline; deployment counts
enabled surfaces across Instagram and Facebook.}
\label{tab:efficiency}\\
\toprule
\textbf{Lever} & \textbf{Metric} & \textbf{Result} \\
\midrule
\endfirsthead
\caption[]{Deployment efficiency and breadth (continued).}\\
\toprule
\textbf{Lever} & \textbf{Metric} & \textbf{Result} \\
\midrule
\endhead
\bottomrule
\endlastfoot
Grouped versus uncompressed history & Prompt tokens & 39.5\% fewer \\
History budgeting versus unbudgeted history & Estimated accelerator requirement & 38.9\% lower \\
Per-string versus per-user annotation & Logical annotation requests & $\approx$98\% fewer \\
Updated versus preceding annotation pipeline & Throughput per accelerator & +145.5\% \\
Parallel versus serial request-time attribution & Chunk-0 server latency & $\approx$5 ms lower \\
Deployment & Enabled surfaces / platforms (Instagram and Facebook) & 4 / 2 \\
Deployment & Time in production & More than one year \\
\end{longtable}
\endgroup

\subsection{Interest-state contract}

Every surface maps its user-facing actions to the same normalized interest identity and explicit-state
precedence. The labels shown in an interface may differ, but the stored states and serving effects below
are shared.

\begingroup
\small
\setlength{\tabcolsep}{3pt}
\renewcommand{\arraystretch}{1.00}
\begin{longtable}{@{}>{\raggedright\arraybackslash}p{0.16\textwidth}
    >{\raggedright\arraybackslash}p{0.25\textwidth}
    >{\raggedright\arraybackslash}p{0.27\textwidth}
    >{\raggedright\arraybackslash}p{0.26\textwidth}@{}}
\caption{\textbf{State and action contract.} Verified state transitions used by profile construction and
serving.}
\label{tab:state-contract}\\
\toprule
\textbf{State} & \textbf{Entry condition} & \textbf{Persistence and precedence} & \textbf{Serving effect} \\
\midrule
\endfirsthead
\caption[]{State and action contract (continued).}\\
\toprule
\textbf{State} & \textbf{Entry condition} & \textbf{Persistence and precedence} & \textbf{Serving effect} \\
\midrule
\endhead
\bottomrule
\endlastfoot
Inferred & Generated from activity and retained after behavioral support and policy checks &
Refreshed with the inferred profile; may expire when support decays &
Available to profile ranking, retrieval, attribution, and display selection \\
Explicit addition & Accepted normalized user text; prior behavioral support is optional &
Stored separately from inference and persists until the user changes it &
Creates an authoritative positive interest identity for topical supply and ranking \\
Explicit promotion & A user promotes an existing normalized identity &
Explicit preference outranks later inferred state &
Raises the priority of matching candidates through the shared identity \\
Explicit removal or suppression & A user removes an entry or requests less of its normalized identity &
Stored as explicit state and not overwritten by later inference &
Excludes the entry from the served profile or applies the corresponding negative serving path \\
\end{longtable}
\endgroup

\subsection{Experiment contracts}

The main tables report effect sizes; Table~\ref{tab:experiment-contracts} collects the corresponding
design and analysis contracts in one place. Details are stated at the level needed to interpret each comparison.

\begingroup
\small
\setlength{\tabcolsep}{2.4pt}
\renewcommand{\arraystretch}{0.94}
\begin{longtable}{@{}>{\raggedright\arraybackslash}p{0.15\textwidth}
    >{\raggedright\arraybackslash}p{0.28\textwidth}
    >{\raggedright\arraybackslash}p{0.25\textwidth}
    >{\raggedright\arraybackslash}p{0.26\textwidth}@{}}
\caption{\textbf{Study contracts for Tables 1--3.} All online component comparisons are randomized by
user. Replenishment and volume-regulation values show relative changes with bracketed uncertainty intervals.}
\label{tab:experiment-contracts}\\
\toprule
\textbf{Study} & \textbf{Treatment and comparison} & \textbf{Analysis population / unit} & \textbf{Endpoint and uncertainty} \\
\midrule
\endfirsthead
\caption[]{Study contracts for Tables 1--3 (continued).}\\
\toprule
\textbf{Study} & \textbf{Treatment and comparison} & \textbf{Analysis population / unit} & \textbf{Endpoint and uncertainty} \\
\midrule
\endhead
\bottomrule
\endlastfoot
Direct review & One deployed displayed profile; descriptive review &
7,261 displayed interests from 393 users; pooled over entries &
Entry rated accurate; descriptive percentage \\
Generator benchmark & Specialized 3B, Gemini 2.5 Pro, and prior generator; shared history format, reference, and judge &
Specialized and prior: 250 users each; Gemini 2.5 Pro: 232 users &
Mean-user F1@10 / P@10 / R@10 / F1@all; reported model means \\
System-level steerability & Deployed-system descriptive measurements & Newly declared interest, eligible user,
and sampled attributed impression &
79.6\% responsiveness, 82.6\% prevalence, and 87.2\% independently judged topicality \\
Instagram Reels reception & Attributed versus other recommendations for the declared-interest cohort &
Delivered impressions and content-control actions & Selected levels and contrasts \\
Instagram Search reference & Attributed recommendations versus search-result videos &
Delivered impressions; common user sample for skips & Relative like and dwell; percentage-point skip contrast \\
Serving-layer stack ablation & Four cumulative removal arms and production control (Appendix~\ref{app:serving-experiments}) &
Eligible positive-interest lifecycle &
Attributed matching delivery same day and within 16 impressions; $p<0.01$ \\
Suppression-aware ranking & Similarity-aware demotion plus production filter versus filter alone &
Suppressed interest in subsequent recommendations &
Reappearance 11.8\% lower same day, 21.8\% at 7 days, 15.3\% at 14 days, and 24.0\% at 28 days; $p<0.01$ \\
Replenishment & Dynamic topic selection versus the prior fixed sourcing-age policy; separate assignment &
Instagram user; active interests for the engaged-interest ratio, and distinct days for delivery &
Active interests with a match watched for at least three seconds: +8.34\% [+8.07\%, +8.61\%];
User days with explicit-interest delivery: +2.43\% [+2.24\%, +2.63\%] \\
Volume regulation & Real-time per-interest accounting and recalibrated volume dampening versus prior daily accounting &
Instagram user; six-day comparison &
Recorded flooding-days metric: 15.4\% lower [13.4\%, 17.4\%] \\
\end{longtable}
\endgroup

\subsection{Governance and state isolation}

Sensitivity and integrity annotations attach once to each normalized interest string and are applied to
both inferred and user-entered interests. When a user-entered interest is rejected by these gates, the
editing interface displays an error. Shared annotations contain no user-specific membership,
behavioral support, or preference state. Those fields remain in per-user serving records, while explicit
preference state is stored separately from model output and retains precedence across refreshes. This
separation lets the system reuse policy annotations without sharing a person's interest state.

\section{Serving Experiment Details}
\label{app:serving-experiments}

\subsection{Serving-layer stack ablation}

A five-arm user-randomized experiment removes serving layers cumulatively from Instagram's production configuration. The first arm disables the early- and late-stage added-interest ranking terms. The second
additionally disables reserved top-slot insertion. The third additionally disables negative filtering,
explicit suppression, and single-topic delivery-gap rules. The fourth additionally sets
interest-conditioned candidate source weights to zero, leaving no layer enabled. Differences between
consecutive arms give the marginal delivery loss at each removal step; the directly estimated outer-arm
contrast compares all layers removed with the production control.

Eligible positive additions must have sufficient post-edit viewing opportunity. The primary endpoint
counts whether at least one automatically attributed video appears on the day of the edit; a secondary
endpoint counts whether one appears within the first 16 qualifying video impressions. All reported randomized serving-component comparisons have $p<0.01$. On the secondary endpoint the three delivery layers each account for
12--13 percentage points. Table~\ref{tab:stack-ablation} reports the
per-layer losses on the primary endpoint.

Two limits apply. The ablated arms are not production configurations, so their rates are not production
rates. The control arm is measured under a narrower endpoint definition than the same-day responsiveness
in Table~\ref{tab:main} and approximates rather than reproduces it.

\subsection{Suppression-aware ranking}

A separate user-randomized experiment compares the production post-retrieval cluster-ID filter with the
same filter plus similarity-aware demotion of ordinary recommendations. Its endpoint is the share of
suppressed interests that still appear in recommended videos at each horizon; lower is better.
Table~\ref{tab:experiment-contracts} reports the same-day and longer-horizon effects.

\subsection{Replenishment}

The comparison covers seven days. The engaged-interest metric is the ratio of interests seen with
at least three seconds of dwell to active interests; its companion counts distinct days on which a user
received at least one explicit-interest impression. The experiment unit is the Instagram user.

The replenishment treatment removes a fixed sourcing-age cutoff and replaces it with a dynamic selection
policy. Among interests eligible for offline sourcing, those past a cold-start window are admitted only
above a recent-view threshold; the most recently declared interests retain reserved capacity, and the
remaining slots are filled by least recent realized exposure rather than at random. The control retains
the prior age-limited policy. Replenishment increases the share of active interests receiving a match watched for at least
three seconds by 8.34\% [8.07\%, 8.61\%] and user days with explicit-interest delivery by 2.43\% [2.24\%, 2.63\%].

\section{Serving and Steering Detail}
\label{app:steering}

\begin{center}
\begin{minipage}[t]{\textwidth}
  \centering
  \includegraphics[width=\textwidth]{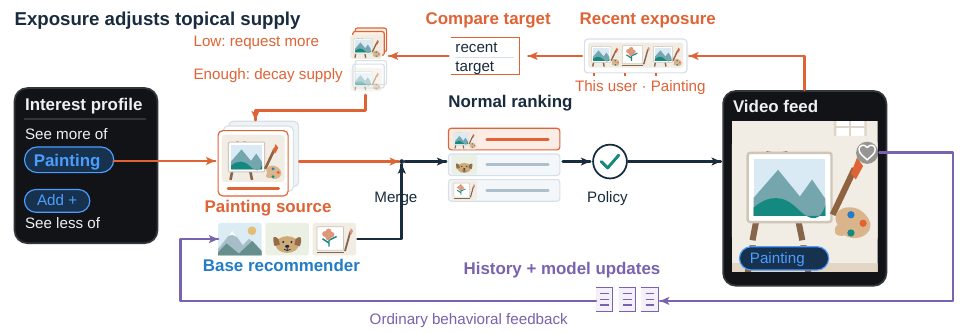}
  \Description{A controller combines topical and ordinary candidates, then uses recent per-interest exposure to request or dampen topical supply; user responses update the recommender through a slower loop.}
  \captionof{figure}{\textbf{Closed-loop delivery control.} A topical source merges with ordinary
  candidates. Stable attribution lets recent exposure request or dampen topical supply; user responses
  return through ordinary history and model updates.}
  \label{fig:controller-detail}
\end{minipage}\par\medskip
\begin{minipage}[t]{\textwidth}
  \centering
  \includegraphics[width=\textwidth]{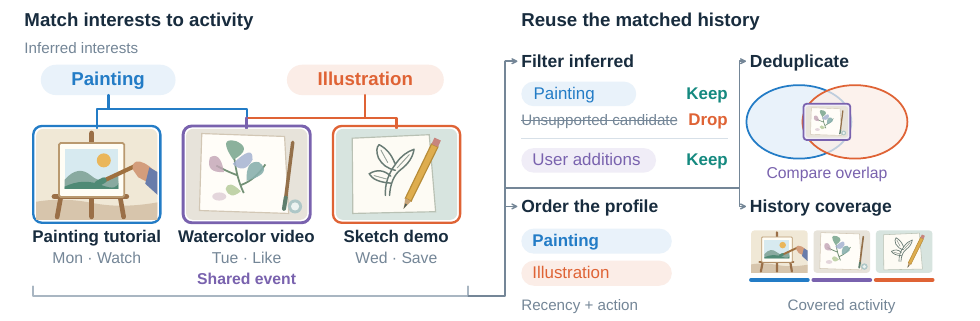}
  \Description{Matched activity supports inferred-interest filtering, personalized deduplication, profile ordering, and history coverage.}
  \captionof{figure}{\textbf{Behavioral grounding and profile-side reuse.} Reverse lookup supplies
  support filtering, personalized deduplication, ordering, and history coverage. Recommendation-side
  consumers appear in Figure~\ref{fig:write-detail}; explicit additions require no prior support.}
  \label{fig:grounding-detail}
\end{minipage}
\end{center}

\FloatBarrier

\end{document}